\documentclass[10pt,journal]{IEEEtran}%
\usepackage{amsfonts}
\usepackage{framed,multirow}
\usepackage{amssymb}
\usepackage{amsmath}
\usepackage{latexsym}
\usepackage{subfigure}
\usepackage{url}
\usepackage{xcolor}
\usepackage{tabularx}
\usepackage{makecell}
\usepackage{multirow}
\usepackage[breaklinks=true,colorlinks,bookmarks=false]{hyperref}
\usepackage{placeins}
\usepackage{booktabs}
\usepackage{graphicx}

\definecolor{newcolor}{rgb}{.8,.349,.1}
\input{tcilatex}
\begin{document}

\thispagestyle{empty}


\title{Attention-Based Multimodal Image Matching}

\author{Aviad Moreshet and Yosi Keller$^{\ast}$
\IEEEcompsocitemizethanks{\IEEEcompsocthanksitem A. Moreshet and Y. Keller are with the Faculty of Engineering, Bar Ilan University, Ramat-Gan, Israel. \protect \and
Email: yosi.keller@gmail.com
}\thanks{Manuscript received April 19, 2005; revised August 26, 2015.}}

\maketitle

\begin{abstract}
We propose a method for matching multimodal image patches using a multiscale Transformer-Encoder
that focuses on the feature maps of a Siamese CNN. It effectively combines multiscale image embeddings while improving task-specific and appearance-invariant image cues. We also introduce a residual attention architecture that allows for end-to-end training by using a residual connection. To the best of our knowledge, this is the first successful use of the Transformer-Encoder architecture in multimodal image matching. We motivate the use of task-specific multimodal descriptors by achieving new state-of-the-art accuracy on both multimodal and unimodal benchmarks, and demonstrate the quantitative and qualitative advantages of our approach over state-of-the-art unimodal image matching methods in
multimodal matching. Our code is shared here: \href{https://github.com/CodeJjang/multiscale-attention-patch-matching}%
{Code}.
\end{abstract}




\section{Introduction}

The matching of image patches is a fundamental task in computer vision,
aiming to determine the correspondence of detected keypoints by encoding and
matching the patches around them using feature point detectors and
descriptors. Multimodal patch matching focuses on matching patches
originating from different sensors, such as visible RGB and near-infrared
(NIR). Such patches are inherently more difficult to match due to the
nonlinear and unknown variations in pixel intensities, varying lighting
conditions, and colors. In particular, multimodal images are manifestations
of different physical attributes of the depicted objects and the acquisition
devices. There is a wide range of multimodal patch matching applications,
such as information fusion in medical imaging devices \cite%
{MedicalImageRegistration} and multisensor image alignment \cite%
{MultiSensorAlignment}.

Pioneering work \cite{SIFT_RT, symmetric_SIFT, multispectral_image_fp}
extended classical local image descriptors such as SIFT \cite{SIFT} and SURF
\cite{surf} to multimodal matching, attempting to derive
appearance-invariant descriptors. Matching the resulting descriptors using
the Euclidean distance yielded limited results, as high-level structural
information could not be captured. Recently, deep learning models have been
shown to be the most effective \cite{PN_net, Q-net, TS-net, multisensor,
BetterAndFaster}, allowing for improved performance in multiple modalities.
Contemporary state-of-the-art (SOTA) approaches are based on training a
Siamese CNN \cite{siamesenn}, either by directly outputting pairwise patch
matching probabilities \cite{Zagoruyko,MatchNet,TS-net}, or by outputting
the patches feature descriptors and encode their similarity in a latent
space \cite{SimoSerra,HardNet,L2Net}, an approach denoted as descriptor
learning.

Different losses were used to optimize the networks. Cross-Entropy Loss was
used \cite{MatchNet, Q-net, TS-net} to classify image patches as a binary
same/not same classification task. Contrastive \cite{SimoSerra, multisensor}
and Triplet Losses \cite{HardNet} were used along with Euclidean distance to
learn descriptors encoding latent space similarities. In recent studies,
hard negative mining schemes \cite{PN_net, HardNet, multisensor, Twin_net}
have become commonplace due to the high ratio of negative pairs, most of
which are easy negatives that barely contribute to the loss. They are
sometimes combined with hard positive mining \cite{SimoSerra, Q-net,
BetterAndFaster}. Detecting and encoding local features is the first step in
the overall matching scheme. In the next step, the local features are used
as input to matching schemes such as the seminal RANSAC\ \cite{RANSAC} and
its extensions \cite{MAGSAC++}, as well as recent deep learning based
approaches: CoAM \cite{CoAM}, LoFTR \cite{LoFTR}, and SuperGlue \textbf{\cite%
{SuperGlue}}, to name a few.
\begin{figure*}[tb]
\centering\includegraphics[width=0.9\textwidth]{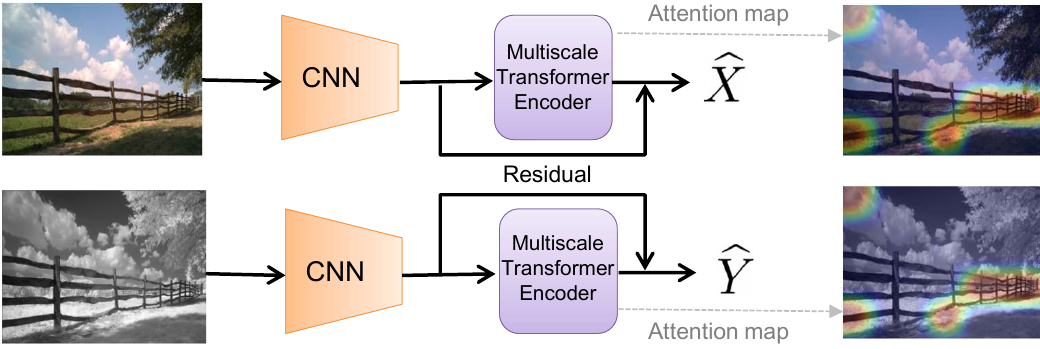}
\caption{The proposed attention-based approach. A Siamese backbone CNN
computes the feature maps of the input multimodal image patches that are
passed through a Spatial Pyramid Pooling (SPP) layer \protect\cite{SPP}. A
Transformer-Encoder aggregates the multiscale image embeddings. A residual
connection bypasses the Encoder to facilitate end-to-end training. The
attention heatmaps visualize the modality-invariant visual cues emphasized
by the attention scheme.}
\label{fig:arch}
\end{figure*}

In this work, we propose a novel approach for attention-based multimodal
patch matching. The gist of our approach, shown in Fig. \ref{fig:arch}, is
to apply attention using a multiscale Transformer-Encoder to aggregate \emph{%
multiscale} image embeddings into a unified image descriptor. Multiscale
feature maps allow the encoder to capture additional similarities between
the multimodal patches. The descriptors are learned for both modalities by a
weight-sharing Siamese network, which simultaneously learns the joint
informative cues in the modalities of both patches. In particular, as
depicted in the attention heatmaps visualized in Figs. \ref{fig:arch} and %
\ref{fig:attention_heatmaps}, the modality-invariant informative visual cues
correspond to the highest attention activations, computed by the encoder.
Other image locations that are less task-informative show weak attention
activations. To apply the Transformer-Encoder, we formulate the matching
task as a sequence-to-one problem, where the inputs are multimodal feature
maps flattened into sequences, and the outputs are adaptively aggregated
embeddings. The spatial layout information of the image embeddings is
induced by positional encoding. Due to the lack of pre-trained CNN backbones
for multimodal patch matching, the backbone CNN has to be trained from
scratch, starting from randomly initialized weights. This has become the de
facto initialization scheme among other models in the task. We show that
random initialization prevents the encoder from learning meaningful sequence
aggregations at the beginning of the training process, hindering overall
learning. To mitigate this issue, we propose adding a residual connection to
the network bypassing the Transformer-Encoder, ultimately allowing
end-to-end training of the proposed architecture from scratch.

Our approach is experimentally shown in Section \ref{sec:experimental} to
achieve the new SOTA multimodal matching accuracy on multiple datasets
compared to the contemporary \textit{multisensor-specific} descriptor
learning schemes. Moreover, its applicability is further motivated by
showing in Section \ref{subsec:matching accuracy}, that when applied with
SOTA \textit{general-purpose} matching schemes, including CoAM \cite{CoAM},
LoFTR \cite{LoFTR} and SuperPoint \cite{SuperPoint}, it outperforms these
schemes in multisensor matching. Our results show that multisensor
descriptor learning benefits from task-specific schemes such as ours, and
motivates their use.

We highlight the following key contributions.

\begin{enumerate}
\item Introduction of a unique attention-based methodology for multimodal
image patch matching. This method aggregates Siamese CNN feature maps using
a multiscale Transformer-Encoder. To our knowledge, this represents the
inaugural application of the Transformer-Encoder architecture to the task of
multimodal image patch matching.

\item Proposal of a residual connection that bypasses the encoder,
facilitating end-to-end training from scratch. This is pivotal given the
common practice of random initialization and the absence of pre-trained
backbones.

\item Demonstrated state-of-the-art (SOTA) performance of our method on
contemporary multimodal image patch matching benchmarks, including VIS-NIR
\cite{SiameseCrossSpectral} and En et al. \cite{Q-net, TS-net}.
Additionally, our method sets new SOTA standards on the single modality UBC
Benchmark \cite{UBC_benchmark}, underscoring the broad applicability of our
approach.

\item Evidence that when our proposed multisensor image descriptors are
combined with leading matching schemes \cite{CoAM,LoFTR,SuperPoint}, they
surpass matching schemes that utilize general-purpose image descriptors
trained on both RGB and multisensor datasets.
\end{enumerate}

\section{Related work}

\subsection{Shallow multisensor image matching}

The matching of multi sensor images has been studied in a gamut of works.
Earlier works, preceding contemporary CNN-based schemes, were based on
unsupervised approaches, mostly based on deriving appearance-invariant image
representations of multisensor images, that utilized salient image edges and
contours. A robust image representation technique grounded in statistical
methods was developed by Viola and Wells~\cite{Viola:1997:AMM:263008.263015}%
. The mutual information between these statistical representations was
fine-tuned to optimize motion parameters. An iterative, implicit similarity
algorithm for global image alignment was devised by Keller et al.~\cite%
{keller2006multisensor}. Leveraging gradient information, this approach
identifies pixels with peak gradients in one image and optimizes the
gradients at corresponding locations in the second image. This is done in
relation to global parametric motion, without the need for explicit
similarity measure optimization. Irani et al. introduced a hierarchical,
coarse-to-fine strategy for estimating global parametric motion between
multimodal images \cite{irani1998robust}. Using directional derivative
magnitudes as a robust form of image representation, the correlation between
these representations was iteratively maximized. This was achieved through
the application of gradient ascent in a coarse-to-fine manner.
Local modulality-invariant descriptors were derived by modifying the SIFT
descriptor~\cite{chen2009real,MI-SIFT}. Contrast-invariance was achieved by
mapping the gradient orientations of interest points from $[0,2\pi )$ to $%
[0,\pi )$. However, Hasan et al. showed that such mappings decrease the accuracy of the matching~\cite{hossain2011improved}. They further modified SIFT~\cite%
{hasan2012modified} by thresholding gradient values to reduce the impact of
strong edges. Enlarging the spatial window with additional sub-windows
improved spatial resolution. Rather than using gradients, Aguilera et al.~%
\cite{aguilera2012multispectral} employed histograms of contours and edges
to avoid SIFT ambiguities in multimodal images. The dominant orientation was
determined similarly. They later extended this approach by incorporating
multi-oriented, multi-scale Log-Gabor filters~\cite{lghd2015}. Keller and
Averbuch introduced an energy minimization approach to register multimodal
images \cite{1191008}. The alignment of the location of the gradient maxima is used as a
robust matching criterion formulated as parametric optimization. The method
iterates using one image's intensity and initializing with the other's
gradient maxima. A coarse-to-fine scheme enables estimating global motion
even with complex spatially-varying intensity transformations. This implicit
matching criterion utilizes full spatial information without needing
invariant representations. The technique is demonstrated by registering real
multi-sensor images using affine and projective motion models. Geometric
structure was used as an appearance-invariant representation by quantifying
similarity through structural similarity measures. Shechtman et al.~\cite%
{shechtman2007matching} introduced Self Similarity Matching (SSM), a
geometric descriptor encoding local structure around feature points. This
was done by correlating a central patch with all adjacent patches within a
radius. Kim et al.~\cite{kim2015dasc} extended SSM with the Dense Adaptive
Self-Correlation (DASC) descriptor. This computed a self-similarity measure
via adaptive self-correlation between randomly sampled patches.

Ye et al. proposed a robust multimodal image registration method \cite%
{YE2022331} to address nonlinear radiometric differences and geometric
distortions between remote sensing images (optical, infrared, LiDAR, SAR). A
novel Steerable Filters descriptor combining first- and second-order
gradient information depicts discriminative structure features using a
multi-scale strategy. Fast Normalized Cross-Correlation similarity measure
using FFT and integral images improves efficiency. A coarse-to-fine system
first conducts local coarse registration using interest points and geometric
correction, utilizing image georeferencing. The proposed descriptor is
robust to radiometric differences in fine registration, detecting control
points by template matching. A robust feature matching method (R$_{2}$FD$%
_{2} $) for multimodal images was proposed by Zhu et. al \cite{10092838}. It
handles both differences in radiation and geometry. A repeatable feature
detector (multichannel autocorrelation of log-Gabor) combines multichannel
autocorrelation and log-Gabor wavelets to detect interest points with high
repeatability and uniform distribution. A rotation-invariant descriptor
(rotation-invariant maximum index map of log-Gabor) includes fast dominant
orientation assignment and feature representation construction. A
rotation-invariant maximum index map addresses rotation deformations. The
descriptor incorporates this rotation-invariant map with the spatial
configuration of DAISY to improve robustness to radiation and rotation
differences. Zhou et al. applied deep learning to refine structure features
and improve image matching \cite{9541389}. The process involves extracting
multi-oriented gradient features to represent the structure properties of
images, then using a shallow pseudo-Siamese network to convolve the gradient
feature maps in a multiscale manner. This results in multiscale
convolutional gradient features (MCGFs), which are used for image matching
via a fast template scheme. MCGFs can capture finer common features between
SAR and optical images than traditional handcrafted structure features, and
can also overcome some limitations of current deep learning-based matching
methods. The performance of MCGFs was evaluated using two sets of SAR and
optical images with different resolutions.

\subsection{CNN-based approaches for image matching}

Deep learning techniques became the staple of computer vision and were
applied to multimodal matching, showing SOTA performance. Jahrer et al.\cite%
{siamesenn} introduced a Siamese CNN network for the metric learning of
feature descriptors, outperforming classic hand-crafted descriptors such as
SIFT \cite{SIFT}. Han et al.\cite{MatchNet} added a metric learning network
on top of a Siamese CNN, achieving SOTA results and also decreasing the size
of the learned descriptors. Ofir et al. \cite{8451640} propose a deep
learning approach to align cross-spectral images using a learned descriptor.
A feature-based method detects Harris corners and matches them using a
patch-metric learned on top of a network trained on CIFAR-10. The approach
achieves accurate sub-pixel alignment of cross-spectral images. Zagoruyko
and Komodakis \cite{Zagoruyko} explored multiple Siamese-based CNN
architectures. The first is a 2-Channel architecture where the patches are
concatenated channel-wise and are fed to a single CNN. The second is a
Pseudo-Siamese CNN network with non-shared weights, and the last is a
central surrounding architecture, in which two Siamese CNNs are jointly
trained with different input resolutions to better capture multiresolution
information. All methods presented promising results on the reported
benchmarks. Aguilera et al.\cite{SiameseCrossSpectral} extended those
architectures to the multimodal case, outperforming previous approaches.
Balntas et al. suggested a weight-sharing triplet CNN termed PN-Net \cite%
{PN_net}, using image triplets consisting of two matching patches in
addition to a non-matching one. A new loss was introduced to replace hard
negative mining, by penalizing small $L_{2}$ distances between non-matching
pairs and large $L_{2}$ distances between matching pair, forcing the network
to always perform backpropagation using the hardest non-matching sample of
each triplet. Aguilera et al. proposed a cross-spectral extension to the
previous work, a weight-sharing quadruple network called Q-Net \cite{Q-net},
utilizing two pairs of cross-spectral patches. The loss was extended to the
quadruple case, mining both hard positive and hard negative samples at the
same time. L2-Net \cite{L2Net} proposed by Tian et al. addressed an inherent
problem in patch matching, where the number of negative samples is orders of
magnitude larger than the positive samples, picking $n^{2}-n$ hard negative
samples per batch with a minimum distance to the $n$ positive batch pairs.
Mishchuk et al. proposed the HardNet extension \cite{HardNet} to L2-Net \cite%
{L2Net} using a hard negative sampling scheme, ensuring that the distance of
the selected negative sample is minimal. Both approaches achieved SOTA
performance on unimodal benchmarks. Wang et al. introduced a novel
Exponential Loss function \cite{BetterAndFaster} forcing the network to
learn more from both positive and negative hard samples than from easy ones.
The scheme combines hard negative and positive mining inspired by Mishchuk
et al.\cite{HardNet} and Simo et al.\cite{SimoSerra}, respectively. Irshad
et al. followed this line with the Twin-Net \cite{Twin_net} approach, by
proposing to mine twin negatives along with a dedicated Quad Loss, designed
for single modality patches. In twin negatives mining, the first negative is
mined as in Mishchuk et al.\cite{HardNet}, and its closest negative is
picked as the second negative. The Quad Loss then forces the descriptors of
the chosen positive and anchor to be closer than the twin negatives
descriptors, resulting in descriptors with a greater discriminatory power.
Zhang and Rusinkiewicz \cite{ZhangCDF} proposed an improved triplet loss
formulation for feature learning, adaptively modifying the hard margin used
to screen the hard negative. The authors suggest replacing the hard margin
with a nonparametric soft margin, which is dynamically updated based on the
cumulative distribution function of the triplets' signed distances to the
decision boundary.

En et al. proposed an approach to utilize both joint and modality-specific
information within the image patches, by combining Siamese and Pseudo
Siamese CNNs in a unified architecture dubbed TS-Net \cite{TS-net}.
Ben-Baruch and Keller \cite{multisensor} introduced a similar hybrid
framework fusing Siamese and Pseudo Siamese CNNs features through fully
connected layers. Contrary to TS-Net \cite{TS-net}, their approach
calculated $L_{2}$-optimized patch descriptors essential for patch matching,
and optimized the network using multiple auxiliary losses. These extensions
have proven vital and have resulted in the performance of SOTA on various
multimodal benchmarks. Quan et al. proposed the AFD-Net framework \cite%
{AFD_net} consisting of three subnetworks, the first aggregates multilevel
feature differences, the second extracts domain-invariant features from
image patch pairs, while the final one infers matching labels. This approach
computes a patch similarity score, rather than image embeddings. Thus, $%
\mathcal{O}(n^{2})$ forward passes of the network are required for matching
two images containing $n$ image patches each, in contrast to $\mathcal{O}(n)$
forward passes in a descriptor learning approach such as ours.

\subsection{Transformer-based approaches in computer vision}

The transformer architecture was introduced by Vaswani et al. \cite%
{Transformer} as a novel formulation of attention-based approaches, which
allows encoding sequences without RNN layers such as LSTM and GRU.
Transformers were first applied in Natural Language Processing tasks \cite%
{bert}, and then proved successful in a variety of computer vision schemes
\cite{AttentionAugmentedConvNets,
detr,ImageWorth16x16Words,ImageTransformer, SuperGlue}. Transformers, in
contrast to convolution networks, are capable of aggregating long-range
interactions between a sequence of input vectors. In computer vision, we can
formulate the outputs of a backbone CNN as a sequence as in \cite%
{detr,SuperGlue,ImageWorth16x16Words} and encode them using a Transformer
encoder that aggregates spatial interactions (attention weights) between the
activation map entries. Task-informative entries are numerically enhanced
compared to noninformative ones. The Transformer architecture is composed of
stacked layers of self-attention encoders and decoders attending the encoder
outputs using a set of vectors denoted as queries. Using self-attention, the
encoder produces a weighted average over the values of its inputs, such that
the weights are produced dynamically using a similarity function between the
inputs. Transformer-based encoders and decoders utilize multiple stacked
multi-head Attention and feedforward layers. In contrast to the sequentially
structured RNNs, the relative position and sequential order of the sequence
elements are induced by positional encodings that are added to the Attention
embeddings.

\subsection{Attention-based approaches for unimodal image matching}

Attention-based approaches have recently been applied in unimodal,
unconstrained image matching frameworks. SuperGlue \cite{SuperGlue} by
Sarlin et al. studied feature matching using a graph neural network in
combination with self- (intra-image) and cross- (inter-image) attention to
leverage both the spatial relationships of the features and their visual
appearance. Local features are extracted by CNN-based methods such as
SuperPoint \cite{SuperPoint}, or traditional descriptors like SIFT \cite%
{SIFT}. The SOLAR framework \cite{SOLAR} by Ng et al. for image retrieval
and descriptor learning utilizes SOSNet \cite{SOSNet} for second-order
similarity as a loss term and also takes advantage of second-order spatial
relations between spatial features.

Wiles et al. introduced a spatial co-attention mechanism coined CoAM \cite%
{CoAM} for determining correspondences between image pairs with large
differences in illumination, viewpoint, context, and material. This approach
conditions on both images to implicitly account for their differences. It
introduces two key components - a spatial co-attention module (CoAM) that
conditions learned features on both images, and a distinctiveness score to
pinpoint the most suitable matches. The LoFTR approach \cite{LoFTR} by Sun
et al. is of particular interest as it introduces an attention and
Transformer-based local image feature matching approach that archived SOTA
matching accuracy in unimodal images. The LoFTR technique initiates by
establishing dense pixel-wise matches between images at a coarse scale,
subsequently refining these matches at a more detailed scale. By employing
self-attention and cross-attention layers within Transformers, LoFTR derives
feature descriptors influenced by both images.

In contrast, we propose to combine a Siamese CNN for feature extraction with
self-attention implemented by a \textit{multiscale} Transformer-Encoder.
while, LoFTR uses a \textit{single} resolution scale at each network layer
overlooking the multiscale structure of multisensor matching that is based
on matching elongated edges and image structures, as shown in Section \ref%
{subsec:Attention maps} and Fig. \ref{fig:attention_heatmaps}. The use of a
single resolution scale is appropriate in RGB images, where the higher scale
(smaller) objects and image structures are captured similarly in all images
up to appearance variations. Moreover, both the LoFTR \cite{LoFTR} and CoAM
\cite{CoAM} use cross-attention to compute image matching rather than
descriptor learning, as in our approach. LoFTR's Image matching is jointly
applied to both images, implying that it computes the matching between two
particular images, while our approach computes image descriptors \textit{%
separately} for each image. This implies that when matching an input image $%
\mathbf{x}$ to a set of $n$ reference images $\left\{ \mathbf{y}_{i}\right\}
_{1}^{n}$, LoFTR has to be applied $n$ times for the tuples $\left\{ \mathbf{%
x,y}_{i}\right\} _{1}^{n}$, while the proposed descriptor learning scheme
has to applied just once to compute the descriptors for $\mathbf{x}$. We
further demonstrate that our model can be trained end-to-end from scratch
using the small training datasets available for multisensor matching, by
using a novel residual connection, that circumvents the multiscale
Transformer-Encoder. This allows the training signal to flow directly form
the loss to the CNN layers in the first few training epochs. Compared to
LoFTR \cite{LoFTR}, CoAM \cite{CoAM} and SuperGlue \cite{SuperGlue}, our
approach studies descriptor learning only and can be applied with any
matching scheme such as RANSAC \cite{RANSAC} and SuperGlue.

\section{Multiscale attention-based multimodal patch matching}

We propose a novel attention-based approach for multimodal image patch
matching by encoding Siamese CNN feature maps using the multiscale
Transformer-Encoder architecture. We also introduce a residual connection
bypassing the encoder, that is crucial for end-to-end training. We model the
task as a sequence-to-one problem, where the inputs are multiscale feature
maps of patches transformed into sequences, and the output is a learned
feature descriptor. An overview of the proposed scheme is shown in Fig. \ref%
{fig:arch}.

Let $\{\mathbf{X},\mathbf{Y}\}\in
\mathbb{R}
^{H\times W\times C}$ be a pair of multimodal image patches. First, we
compute the corresponding activation maps $\{\overline{\mathbf{X}},\overline{%
\mathbf{Y}}\}\in \mathbb{R}^{\overline{H}\times \overline{W}\times \overline{%
C}}$ using the Siamese CNN backbone. A spatial pyramid pooling layer (SPP)
\cite{SPP} is applied to $\{\overline{\mathbf{X}},\overline{\mathbf{Y}}\}$,
to calculate a multiscale activation map $\{\mathbf{\widetilde{X}}_{k},%
\mathbf{\widetilde{Y}}_{k}\}\in \mathbb{R}^{\widetilde{H_{k}}\times
\widetilde{W}_{k}\times \overline{C}}$, $k=1..K,$ as detailed in Section \ref%
{subsec:backbone}. Each of $\{\mathbf{\widetilde{X}}_{k},\mathbf{\widetilde{Y%
}}_{k}\}$ is flattened into a sequence and aggregated into a single vector
by the Transformer-Encoder, as in Section \ref{subsec:multiscale_transformer}%
. A fully connected layer finally maps the output representations into $\{%
\mathbf{\widehat{X}},\mathbf{\widehat{Y}}\}\in
\mathbb{R}
^{128}$. We also propose a novel residual connection described in Section %
\ref{subsec:residual_connection} that bypasses the Transformer-Encoder and
allows end-to-end training. The network is trained end-to-end using triplet
loss \cite{FaceNetTripletLoss} and a symmetric formulation of the HardNet
approach \cite{HardNet} detailed in Section \ref{subsec:symmetric triplett}.

\subsection{Backbone CNN}

\label{subsec:backbone}

For every pair of multimodal image patches $\{\mathbf{X},\mathbf{Y}\}$, we
apply a Siamese CNN to extract the feature maps $\{\overline{\mathbf{X}},%
\overline{\mathbf{Y}}\}$. The CNN architecture is described in Table \ref%
{tab:cnn}. We use the same backbones as previous works \cite%
{BetterAndFaster, Zagoruyko}, to experimentally exemplify the efficiency of
the proposed attention-based scheme in Section \ref{sec:experimental}. Each
convolutional layer is followed by Batch Normalization and ReLU activation.
We use an SPP layer \cite{SPP} consisting of a four-level pyramid pooling,
transforming the CNN feature maps into $K$ multiscale feature maps $\{%
\mathbf{\widetilde{X}}_{k},\mathbf{\widetilde{Y}}_{k}\}\in \mathbb{R}^{%
\widetilde{H_{k}}\times \widetilde{W}_{k}\times \overline{C}}$ such that $%
\mathbf{\widetilde{H}}_{k}$ and $\mathbf{\widetilde{W}}_{k}\in \{8,4,2,1\}$.
\begin{table}[tbh]
\setlength\tabcolsep{4pt} \centering
\begin{tabular}{cccccc}
\toprule Layer & Output & Kernel & Stride & Pad & Dilation \\
\midrule Conv0 & $64\times 64\times 32$ & $3\times 3$ & 1 & 1 & 1 \\
Conv1 & $64\times 64\times 32$ & $3\times 3$ & 1 & 1 & 1 \\
Conv2 & $31\times 31\times 64$ & $3\times 3$ & 2 & 1 & 2 \\
Conv3 & $31\times 31\times 64$ & $3\times 3$ & 1 & 1 & 1 \\
Conv4 & $29\times 29\times 128$ & $3\times 3$ & 1 & 1 & 2 \\
Conv5 & $29\times 29\times 128$ & $3\times 3$ & 1 & 1 & 1 \\
Conv6 & $29\times 29\times 128$ & $3\times 3$ & 1 & 1 & 1 \\
Conv7 & $29\times 29\times 128$ & $3\times 3$ & 1 & 1 & 1 \\
\bottomrule &  &  &  &  &
\end{tabular}%
\caption{The architecture of the Siamese CNN backbone.}
\label{tab:cnn}
\end{table}

\subsection{Multiscale Transformer-Encoder}

\label{subsec:multiscale_transformer}

\begin{figure}[t]
\caption{The proposed multiscale Transformer-Encoder architecture. $%
\overline{\mathbf{X}}$ is a feature map computed by the backbone CNN, and
pooled into four multiscale feature maps $\mathbf{\widetilde{X}}_{k}$ by a
SPP layer \protect\cite{SPP}. The Transformer-Encoder aggregates the pooled
maps $\mathbf{\widetilde{X}}_{k}$ along with their positional encodings, and
the aggregations $\mathbf{O}_{k}^{X}\in
\mathbb{R}
^{128}$ are concatenated alongside the residual connection.}
\label{fig:arch_detailed}\centering\includegraphics[width=1.0%
\linewidth]{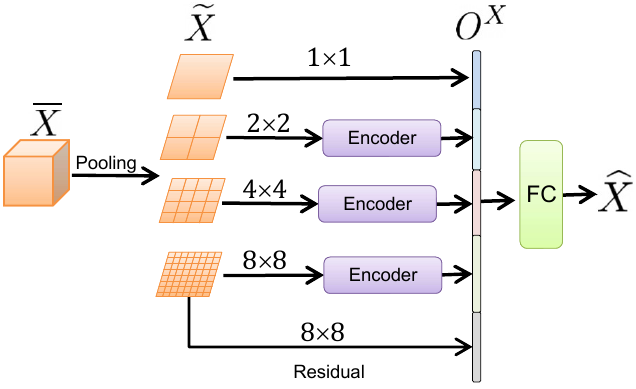}
\end{figure}

The proposed \emph{weight-sharing} multiscale Transformer-Encoder aggregates
the $K$ multiscale outputs of the backbone CNN and SPP layer, and is
illustrated in Fig. \ref{fig:arch_detailed}. The encoder has two layers,
each consisting of two multi-head attention blocks. Given $K$ multiscale
feature maps $\{\mathbf{\widetilde{X}}_{k},\mathbf{\widetilde{Y}}_{k}\}$,
each map is spatially flattened into a sequence of size $\widetilde{H}%
_{k}\ast \widetilde{W}_{k}\times \overline{C}$. \ Since the last SPP\ layer $%
\{\mathbf{\widetilde{X}}_{4},\mathbf{\widetilde{Y}}_{4}\}$ is of dimensions $%
\in \mathbb{R}^{1\times \overline{C}}$ there is no need for additional
spatial pooling, and it is passed as is, and concatenated in the output
vecror.

We assign a learnable embedding to each sequence whose corresponding output
is the aggregated result, the same as \emph{[CLS] token used in BERT} \cite%
{bert}. In order to induce spatial information to the permutation-invariant
Transformer encoder, we learn separable 2D positional encodings. The
encoding is learned separately for the $X$ and $Y$ axes to reduce the number
of learned parameters and are denoted as $\mathbf{E}_{X}\in \mathbb{R}^{%
\widetilde{H}_{k}\times \frac{\overline{C}}{2}}$ and $\mathbf{E}_{Y}\in
\mathbb{R}^{\widetilde{W}_{k}\times \frac{\overline{C}}{2}}$ respectively.
The final 2D positional encoding $\mathbf{E}_{i,j}\in \mathbb{R}^{\overline{C%
}}$ is given by:%
\begin{equation}
\mathbf{E}_{i,j}=%
\begin{bmatrix}
\mathbf{E}_{X}^{j} \\
\mathbf{E}_{Y}^{i}%
\end{bmatrix}%
,\left[ i,j\right] \in \left[ 1...\widetilde{H_{k}},1...\widetilde{W}_{k}%
\right]  \label{eq:2d_emb}
\end{equation}%
Positional encodings are then spatially flattened into a sequence of size $%
\widetilde{H}_{k}\ast \widetilde{W}_{k}\times \overline{C}$ and summed with
the feature map sequences $\{\mathbf{\widetilde{X}}_{k},\mathbf{\widetilde{Y}%
}_{k}\}$. The Transformer-Encoder is then applied to the enriched feature
map sequences to produce $K$ aggregations, denoted $\left\{ \mathbf{O}%
_{k}^{X},\mathbf{O}_{k}^{Y}\right\} $. $\left\{ \mathbf{O}_{k}^{X}\right\} $
and $\left\{ \mathbf{O}_{k}^{Y}\right\} $ are concatenated separately into
two vectors of dimension $\mathbb{R}^{4\overline{C}}$, together with the
feature map of the residual connections $\left\{ \mathbf{\widetilde{X}}_{8},%
\mathbf{\widetilde{Y}}_{8}\right\} \in \mathbb{R}^{8\times 8\times \overline{%
C}}$, as discussed in Section \ref{subsec:residual_connection}. Then each of
the two concatenated embeddings is passed through an FC layer and $L_{2}$
normalized.

\subsection{The residual connection}

\label{subsec:residual_connection}

It is common to apply attention layers to pre-trained backbone CNNs that are
kept frozen during the training process, allowing the Transformer-Encoder to
train based on meaningful feature representations computed by the backbone.
In contrast, in multimodal matching, the backbone has to be trained
end-to-end from scratch, due to the absence of pre-trained weights for the
task. Thus, at the beginning of the training, until the CNN learns to
extract useful feature maps, the encoder does not manage to learn meaningful
sequence aggregations either, hindering the overall learning. Hence, we
propose to resolve this issue by passing the pooled feature map $\left\{
\mathbf{\widetilde{X}}_{8},\mathbf{\widetilde{Y}}_{8}\right\} \in \mathbb{R}%
^{8\times 8\times \overline{C}}$, having the largest spatial support,
directly to the output layer using a residual connection. The residual
connection circumvents the encoder, adding a vital learning signal allowing
end-to-end training from scratch. The residual feature map is flattened into
a vector $\in \mathbb{R}^{8\cdot 8\cdot \overline{C}}$ and concatenated into
the output vector. Note that in practice, the output of the SPP\ layer $\{%
\mathbf{\widetilde{X}}_{4},\mathbf{\widetilde{Y}}_{4}\}$ $\in \mathbb{R}%
^{1\times \overline{C}}$ is also a residual connection (Fig. \ref%
{fig:arch_detailed}), but due to the intensive pooling, down to $1\times 1,$
the embedding is too diluted and does not allow convergence. Thus, as
exemplified in Table \ref{table:ablation}, our architecture cannot be
trained from scratch without the proposed residual connection. Finally, the
concatenated output is mapped by a fully connected layer $FC$ that produces
feature descriptors for each pair of patches, denoted $\{\mathbf{\widehat{X}}%
,\mathbf{\widehat{Y}}\}\in \mathbb{R}^{128}$.

\subsection{Symmetric triplet loss}

\label{subsec:symmetric triplett}

The proposed network is trained by applying the Triplet Loss \cite%
{FaceNetTripletLoss} to the outputs $\{\mathbf{\widehat{X}},\mathbf{\widehat{%
Y}}\}$:%
\begin{equation}
L_{t}\left( a_{i},p_{j},n_{k}\right)
=\sum_{n,j,k}^{N}max(0,m+d(a_{i},p_{j})-d(a_{i},n_{k}))
\label{eq:triplet_loss}
\end{equation}%
where $a_{i},p_{j}$ and $n_{k}$ are triplets of the anchor, positive, and
negative samples, respectively. $m$ is a margin, and $d$ is the Euclidean
distance. Given a training batch of $B$ embeddings $\{\mathbf{\widehat{X}}%
_{i},\mathbf{\widehat{Y}}_{i}\}_{1}^{B}$, for each entry $i$ we have $a_{i}=%
\mathbf{\widehat{X}}_{i},$ $p_{j}=\mathbf{\widehat{Y}}_{i}$ and $n_{k}$ and
is a sample $\mathbf{\widehat{Y}}_{j}\in \{\mathbf{\widehat{Y}}%
_{i}\}_{1}^{B},j\neq i$. Initially, $\mathbf{\widehat{Y}}_{j}$ is chosen
randomly, and then we switch to hard mining. As the hard negative $n_{k}$ is
dynamically chosen within each batch following the HardNet approach \cite%
{HardNet}, we utilize the symmetry between the channels $\mathbf{\widehat{X}}%
_{i}$ and $\mathbf{\widehat{Y}}_{i}$ and train the network using a
symmetrized formulation of Eq. \ref{eq:triplet_loss} as follows:%
\begin{equation}
L_{t}^{s}\left( a_{i},p_{j},n_{k},n_{k^{\prime }}\right) =L_{t}\left(
a_{i},p_{j},n_{k}\right) +L_{t}\left( p_{j},a_{i},n_{k^{\prime }}\right)
\label{eq:symm triplet_loss}
\end{equation}

We used $m=1$ which is a common choice for the margin that also worked well
in previous works. We also experimented using other values, but the
performance did not improve. We chose $d$ as the Euclidean distance function
to be able to utilize efficient large-scale K nearest-neighbor search
schemes for matching descriptors in $L_{2}$ spaces \cite{LSH}.

\section{Experimental Results}

\label{sec:experimental}

We evaluated the proposed scheme using multiple contemporary multimodal
patch matching datasets. We use their publicly available setups that provide
a \textit{deterministic }generation of matching and non-matching train and
validation patches. This allows us to apply a repeatable experimental setup
that is shared and utilized by us and the SOTA schemes against which we
compare. We also demonstrate the generality of our proposal by evaluating it
on the UBC benchmark \cite{UBC_benchmark}, a general single-modality
benchmark for patch matching that was used in previous work. We measure our
performance using the false positive rate at 95\% recall (FPR95), where a
smaller FPR95 score is better.

\subsection{Datasets}

The VIS-NIR Benchmark \cite{SiameseCrossSpectral} is a cross-spectral image
patch matching benchmark consisting of over 1.6 million RGB and
Near-Infrared (NIR) patch pairs of size $64\times 64$. The benchmark was
collected by Aguilera et al. \cite{SiameseCrossSpectral, Q-net} from the
public VIS-NIR scene dataset. Half of the patches are matching pairs, and
half are non-matching pairs chosen at random. A public domain code\footnote{%
VIS-NIR benchmark public setup code: https://github.com/ngunsu/lcsis} by
Aguilera et al. allows to deterministically recreate the training and test
sets, and was used by all schemes we compare against. The En et al.
benchmark \cite{TS-net} consists of three different multimodal datasets
sampled on a uniform grid layout, with a corresponding public domain code%
\footnote{%
En et al. benchmark setup code: https://github.com/ensv/TS-Net}: VeDAI \cite%
{Vedai}, consisting of vehicles in aerial imagery, CUHK \cite{CUHK},
consisting of faces and face sketch pairs, and VIS-NIR \cite%
{SiameseCrossSpectral}. Following the setup of En et al. \cite{TS-net}, 70\%
of the data is used for training, 10\% for validation and 20\% for testing.

The UBC benchmark \cite{UBC_benchmark}, also known as the Brown benchmark,
is a single-mode data set consisting of image pairs sampled from 3D
reconstructions. The benchmark consists of three subsets: Liberty,
Notredame, and Yosemite, consisting of 450K, 468K, and 634K patches,
respectively. Patches were extracted using Difference of Gaussian \cite{SIFT}
or Harris corner detector \cite{harris}. Half of the patches, with the same
3D point, are matching pairs, and the rest are non-matching pairs. Following
\cite{UBC_benchmark, L2Net, Zagoruyko, HardNet} we iteratively train on one
subset and test on 100k pairs of the other subsets.

\subsection{Training}

We use PyTorch \cite{PyTorch} with three NVIDIA GeForce GTX 1080 Ti GPUs.
The networks are randomly initialized from a normal distribution and
optimized using Adam \cite{Adam} with an initial learning rate of 0.1. A
warm-up learning rate scheduler \cite{ImageNetOneHour} was used for eight
epochs and then replaced by a scheduler that reduced the learning rate by a
factor of 10 for every plateau of three epochs. The architecture was trained
for 70 epochs using batches of size 48. The patches were mean-centered and
normalized, and basic augmentations of horizontal flips and rotations were
applied. The architecture is trained with random negatives until there is no
loss decrease for three epochs. We then start mining hard negative samples
using Mishchuk et al. technique \cite{HardNet}. Our code is shared here:
\href{https://github.com/CodeJjang/multiscale-attention-patch-matching}{Code}%
.

\subsection{VIS-NIR Benchmark}

\begin{table*}[tbh]
\centering
\par
\begin{tabular}{lccccccccc}
\toprule Method & Field & Forest & Indoor & Mountain & Old building & Street
& Urban & Water & Mean \\ \midrule
\multicolumn{10}{c}{Handcrafted descriptor methods}  \\ 
SIFT \cite{SIFT} & 39.44 & 11.39 & 10.13 & 28.63 & 19.69 & 31.14 & 10.85 &
40.33 & 23.95 \\
MI-SIFT \cite{MI_SIFT} & 34.01 & 22.75 & 12.77 & 22.05 & 15.99 & 25.24 &
17.44 & 32.33 & 24.42 \\
LGHD \cite{LGHD} & 16.52 & 3.78 & 7.91 & 10.66 & 7.91 & 6.55 & 7.21 & 12.76
& 9.16 \\
\midrule
\multicolumn{10}{c}{Learning-based methods} \\
Siamese \cite{SiameseCrossSpectral} & 15.79 & 10.76 & 11.6 & 11.15 & 5.27 &
7.51 & 4.6 & 10.21 & 9.61 \\
Pseudo Siamese \cite{SiameseCrossSpectral} & 17.01 & 9.82 & 11.17 & 11.86 &
6.75 & 8.25 & 5.65 & 12.04 & 10.32 \\
2-Channel \cite{SiameseCrossSpectral} & 9.96 & 0.12 & 4.4 & 8.89 & 2.3 & 2.18
& 1.58 & 6.4 & 4.47 \\
PN-Net \cite{PN_net} & 20.09 & 3.27 & 6.36 & 11.53 & 5.19 & 5.62 & 3.31 &
10.72 & 8.26 \\
Q-Net \cite{Q-net} & 17.01 & 2.70 & 6.16 & 9.61 & 4.61 & 3.99 & 2.83 & 8.44
& 6.86 \\
TS-Net \cite{TS-net} & 25.45 & 31.44 & 33.96 & 21.46 & 22.82 & 21.09 & 21.9
& 21.02 & 24.89 \\
SCFDM \cite{SCFDM} & 7.91 & 0.87 & 3.93 & 5.07 & 2.27 & 2.22 & 0.85 & 4.75 &
3.48 \\
L2-Net \cite{L2Net} & 16.77 & 0.76 & 2.07 & 5.98 & 1.89 & 2.83 & \textbf{0.62%
} & 11.11 & 5.25 \\
HardNet \cite{HardNet} & 10.89 & 0.22 & 1.87 & 3.09 & 1.32 & 1.30 & 1.19 &
2.54 & 2.80 \\
Exp-TLoss \cite{BetterAndFaster} & 5.55 & 0.24 & 2.30 & 1.51 & 1.45 & 2.15 &
1.44 & 1.95 & 2.07 \\
D-Hybrid-CL \cite{multisensor} & 4.4 & 0.20 & 2.48 & 1.50 & 1.19 & 1.93 &
0.78 & \textbf{1.56} & 1.7 \\
\textbf{Ours} & \textbf{4.22} & \textbf{0.13} & \textbf{1.48} & \textbf{1.03}
& \textbf{1.06} & \textbf{1.03} & 0.9 & 1.9 & \textbf{1.44} \\
\bottomrule &  &  &  &  &  &  &  &  &
\end{tabular}%
\caption{Patch matching results evaluated on the VIS-NIR Benchmark
\protect\cite{SiameseCrossSpectral}. The score is given in terms of FPR95.}
\label{tab:visnir}
\end{table*}

To demonstrate the effectiveness of our approach in the multimodal image
patch matching task, we compare it with 14 state-of-the-art methods, using
the same setup as Aguilera et al. \cite{SiameseCrossSpectral} and \cite%
{TS-net, Q-net, multisensor}. The Country category is divided into 80\% and
20\% for training and validation, respectively. The rest of the categories
are used for testing. The results are shown in Table \ref{tab:visnir}. We
compare our proposal to both handcrafted descriptors \cite{SIFT, MI_SIFT,
LGHD} and state-of-the-art deep learning methods. We quote the results
reported by \cite{BetterAndFaster, multisensor} on this benchmark.

For the handcrafted descriptors \cite{SIFT, MI_SIFT, LGHD} we used their
publicly available code. As in previous works, such approaches are
outperformed by learning-based methods due to their inability to capture
high-level semantic information. PN-Net \cite{PN_net}, L2-Net \cite{L2Net},
HardNet \cite{HardNet} and Exp-TLoss \cite{BetterAndFaster} are single-mode
general-purpose schemes that were applied to the VIS-NIR benchmark by Wang
et al.\cite{BetterAndFaster}. These approaches, which are based on a Siamese
CNN architecture, the same as the proposed scheme, generalized well to the
multimodal case. Aguilera et al.'s three proposals \cite%
{SiameseCrossSpectral} of vanilla Siamese CNN, Pseudo Siamese CNN, 2-Channel
CNN, as well as SCFDM \cite{SCFDM} and D-Hybrid-CL \cite{multisensor} were
specifically designed for multimodal image patch matching and were tested on
the VIS-NIR benchmark. The SOLAR framework training code for local
descriptor learning \cite{SOLAR} was not made public, and the publicly
available pre-trained model performed poorly in the multisensor dataset. So
we avoided reporting its results due to lack of insights. Our proposed
scheme outperforms all previous approaches, achieving a new SOTA on this
benchmark. In particular, it surpasses the previous SOTA Exp-TLoss \cite%
{BetterAndFaster} by 43\% and D-Hybrid-CL \cite{multisensor} by 18\%. Both
methods are based on \textit{the same} backbone CNN and SPP layer.
Therefore, we attribute the improved performance to the proposed
attention-based inference, compared to previous schemes \cite%
{BetterAndFaster,multisensor} that are limited to multiscale SPP-based
inference without spatial attention-based aggregation.

\subsection{En et al. Benchmark}

We apply our approach to this multimodal benchmark following En et al.'s
\cite{TS-net} publicly available setup \footnote{%
En et al. benchmark setup code: https://github.com/ensv/TS-Net} and compare
with ten state-of-the-art methods, both handcrafted \cite{SIFT, MI_SIFT,
LGHD} and learning-based \cite{SiameseCrossSpectral, Q-net, TS-net,
multisensor}. The results are reported in Table \ref{tab:en_etal}. We quote
the results of previous schemes reported by \cite{TS-net, multisensor} on
this benchmark.

On the VeDAI dataset, both 2-Channel \cite{SiameseCrossSpectral}, Hybrid
\cite{multisensor} variants and our approach achieve zero error. On CUHK,
our approach achieves the same performance as Hybrid-CE \cite{multisensor}.
CUHK is a dataset composed of pairs of faces along with their face sketches.
In this case, where the features are relatively limited and spatially close
to each other, it seems that self-attention over the feature maps might not
capture additional information. However, we established a new SOTA in the
VIS-NIR data set, improving it by 93\%.
\begin{table*}[tbh]
\centering
\par
\begin{tabular}{lccc}
\toprule Method & VeDAI & CUHK & VIS-NIR \\
\midrule
\multicolumn{4}{c}{Handcrafted descriptor methods} \\
SIFT\cite{SIFT} & 42.74 & 5.87 & 32.53 \\
MI-SIFT \cite{MI_SIFT} & 11.33 & 7.34 & 27.71 \\
LGHD\cite{LGHD} & 1.31 & 0.65 & 10.76 \\
\midrule
\multicolumn{4}{c}{Learning-based methods} \\
Siamese\cite{SiameseCrossSpectral} & 0.84 & 3.38 & 13.17 \\
Pseudo Siamese\cite{SiameseCrossSpectral} & 1.37 & 3.7 & 15.6 \\
2-Channel\cite{SiameseCrossSpectral} & \textbf{0} & 0.39 & 11.32 \\
Q-Net\cite{Q-net} & 0.78 & 0.9 & 22.5 \\
TS-Net\cite{TS-net} & 0.45 & 2.77 & 11.86 \\
Hybrid-CE \cite{multisensor} & \textbf{0} & \textbf{0.05} & 3.66 \\
Hybrid-CL \cite{multisensor} & \textbf{0} & 0.1 & 3.41 \\
\textbf{Ours} & \multicolumn{1}{l}{\textbf{0}, $\sigma =0$} & \textbf{0.05}
& \textbf{1.76} \\
\textbf{Ours FPR99} & \multicolumn{1}{l}{\textbf{0}, $\sigma =0$} & - & - \\
\bottomrule &  &  &
\end{tabular}%
\caption{Patch matching results evaluated on En et al. Benchmark
\protect\cite{TS-net} consisting of three multimodal datasets sampled on a
uniform grid layout. The score is given in terms of FPR95. On the VeDAI
dataset, we performed a stricter evaluation using FPR99 and reported the
standard deviation out of 10 tests.}
\label{tab:en_etal}
\end{table*}

\subsection{UBC Benchmark}

\begin{table*}[th]
\centering
\begin{tabular}{lcccccccc}
\toprule Method & Training & NOT & YOS & LIB & YOS & LIB & NOT & Mean \\
\multicolumn{1}{c}{} & Testing & \multicolumn{2}{c}{LIB} &
\multicolumn{2}{c}{NOT} & \multicolumn{2}{c}{YOS} &  \\
\midrule SIFT \cite{SIFT} &  & \multicolumn{2}{c}{29.84} &
\multicolumn{2}{c}{22.53} & \multicolumn{2}{c}{27.29} &  \\
MatchNet \cite{MatchNet} &  & 7.04 & 11.47 & 3.82 & 5.65 & 11.60 & 8.70 &
8.05 \\
TL+AS+GOR \cite{SpreadOutDescriptors} &  & 1.95 & 5.40 & 4.80 & 5.15 & 6.45
& 2.38 & 4.36 \\
L2-Net \cite{L2Net} &  & 3.64 & 5.29 & 1.15 & 1.62 & 4.43 & 3.30 & 3.23 \\
CS L2-Net \cite{L2Net} &  & 2.55 & 4.24 & 0.87 & 1.39 & 3.81 & 2.84 & 2.61
\\
L2-Net \cite{L2Net} &  & 2.36 & 4.70 & 0.72 & 1.29 & 2.57 & 1.71 & 2.22 \\
CS L2-Net \cite{L2Net} &  & 1.71 & 3.87 & 0.56 & 1.09 & 2.07 & 1.30 & 1.76
\\
Scale aware \cite{ScaleAware} &  & 0.68 & 2.51 & 1.79 & 1.64 & 2.96 & 1.02 &
1.64 \\
HardNet \cite{HardNet} &  & 0.53 & 1.96 & 1.49 & 1.84 & 2.51 & 0.78 & 1.51
\\
Twin-Net \cite{Twin_net} &  & 1.19 & 2.12 & \textbf{0.41} & \textbf{0.69} &
\textbf{1.58} & 1.62 & 1.27 \\
Exp-TLoss \cite{BetterAndFaster} &  & 0.47 & 1.32 & 1.16 & 1.10 & 2.01 & 0.67
& 1.12 \\
\textbf{Ours} &  & \textbf{0.35} & \textbf{0.91} & 1.31 & 0.85 & \textbf{1.58%
} & \textbf{0.41} & \textbf{0.9} \\
\bottomrule &  &  &  &  &  &  &  &
\end{tabular}%
\caption{Patch matching results evaluated on the UBC Benchmark \protect\cite%
{UBC_benchmark}. Score is given in terms of FPR95.\newline
LIB: Liberty, NOT: Notredame, YOS: Yosemite.}
\label{tab:ubc}
\end{table*}
To illustrate the general applicability of our approach, we compare it with
11 state-of-the-art methods using the popular single modality UBC Benchmark
\cite{UBC_benchmark}, and the publicly available setup and evaluation code
\footnote{%
UBC benchmark setup and evaluation code:
https://github.com/osdf/datasets/tree/master/patchdata}, that was used by
all the schemes we compare against. The results are presented in Table \ref%
{tab:ubc}. We quote the previous results reported on this benchmark by \cite%
{BetterAndFaster, Twin_net}.

Our approach achieved a new SOTA performance when trained on the Liberty and
Yosemite datasets, surpassing the previous SOTA by 24\%. When trained on
Notredame our approach is slightly outperformed by CS L2-Net \cite{L2Net}
and Twin-Net \cite{Twin_net}. We suspect that it is due to the significant
geometric deformations in Notredame, degrading the feature maps extracted by
our Siamese CNN. Notredame also consists of some extremely hard samples that
Twin-Net \cite{Twin_net} might be able to mine better using its improved
hard samples mining scheme.

\subsection{Keypoints matching accuracy}

\label{subsec:matching accuracy}

\begin{figure}[th]
\centering
\begin{tabular}{c}
\subfigure[SuperPoint]{\includegraphics[width=0.8%
\linewidth]{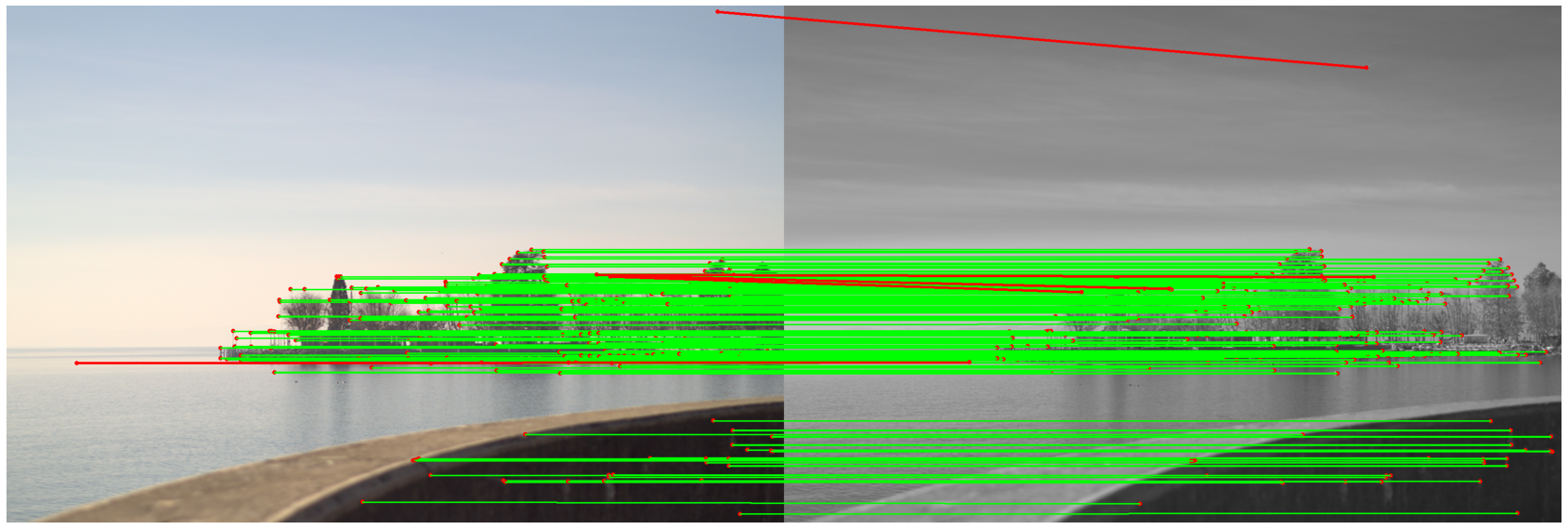}} \\
\subfigure[SuperPoint+Ours]{\includegraphics[width=0.8%
\linewidth]{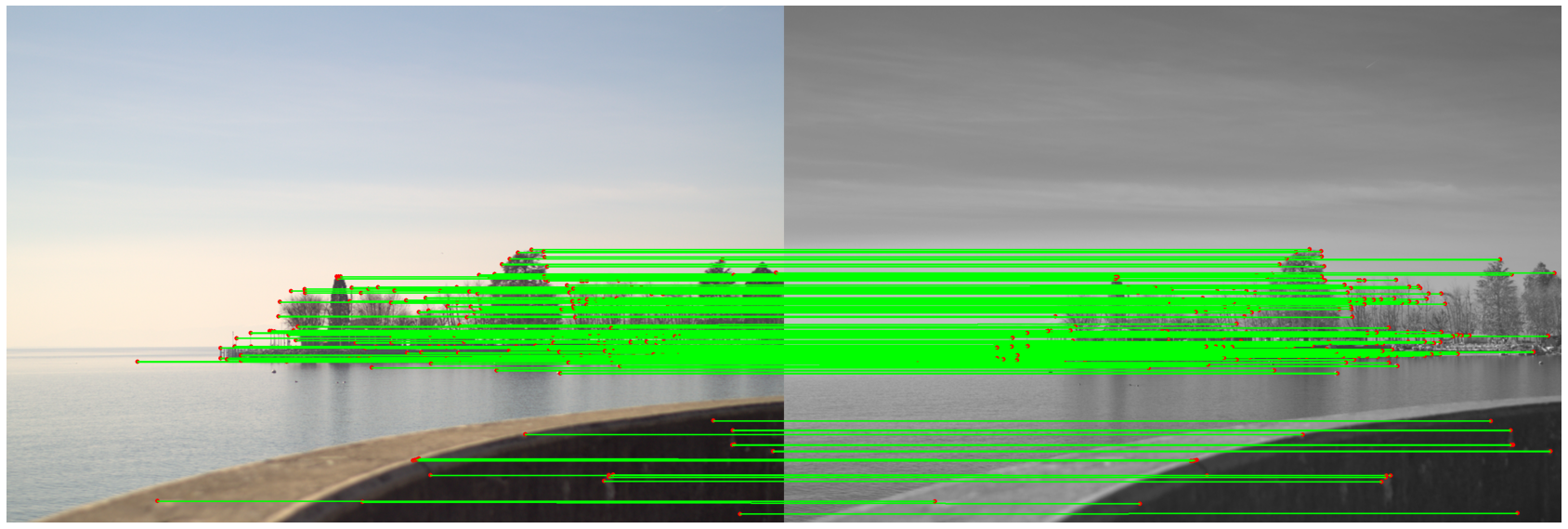}} \\
\subfigure[CoAM]{\includegraphics[width=0.8%
\linewidth]{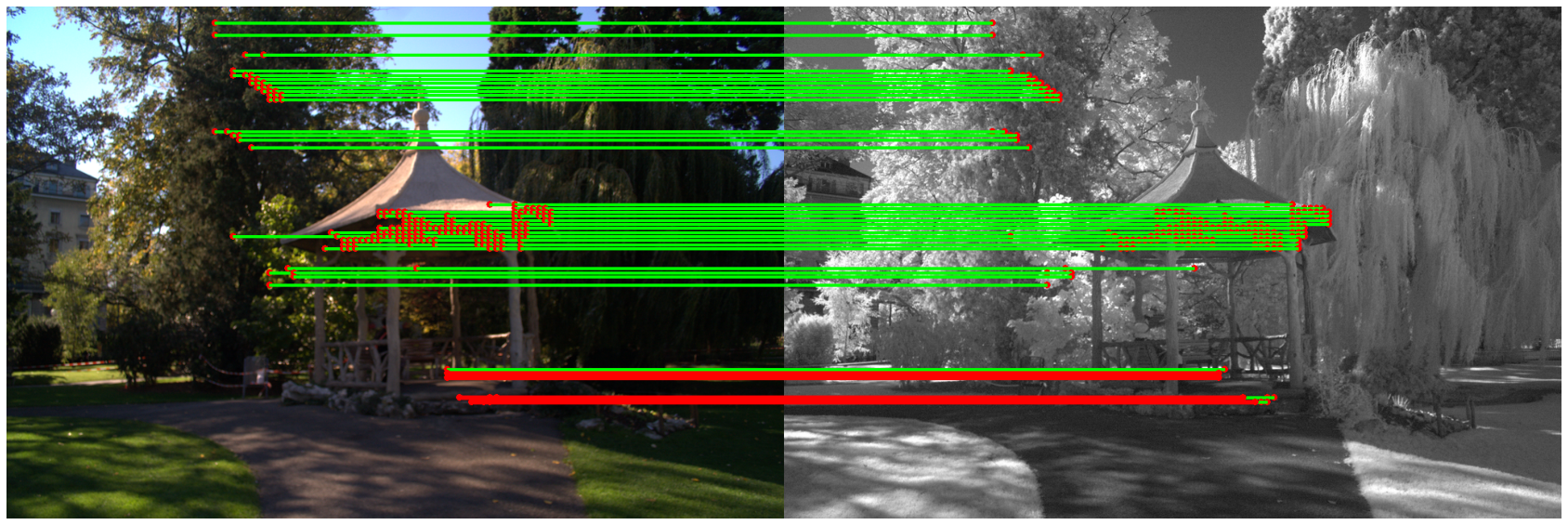}} \\
\subfigure[CoAM+Ours]{\includegraphics[width=0.8%
\linewidth]{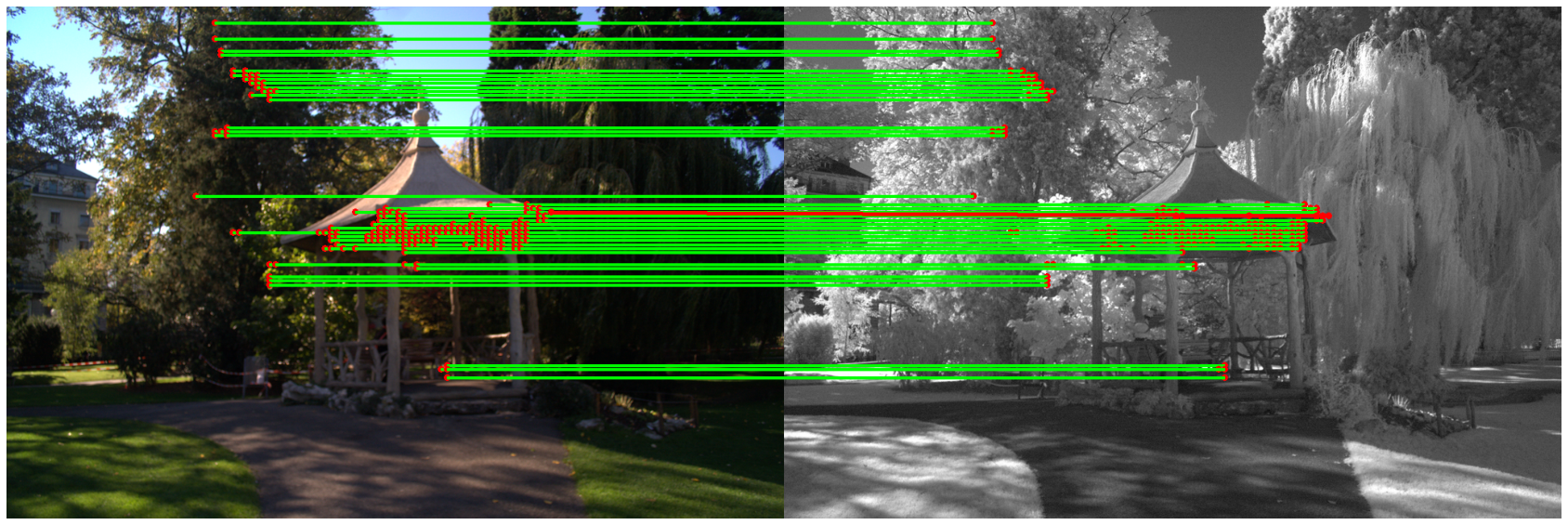}}%
\end{tabular}%
\caption{Keypoints matching visualizations of SuperPoint \protect\cite%
{SuperPoint} in \protect\ref{fig:image_matching}a-\protect\ref%
{fig:image_matching}b and CoAM \protect\cite{CoAM} in \protect\ref%
{fig:image_matching}c-\protect\ref{fig:image_matching}d with and without
using the proposed scheme, on the VIS-NIR Benchmark \protect\cite%
{SiameseCrossSpectral}. Inliers are marked in green and outliers in red. Our
method reduces the number of outliers and produces more accurate results.}
\label{fig:image_matching}
\end{figure}

The proposed multimodal descriptor can also serve as a more accurate drop-in
replacement for multimodal keypoint description in contemporary image
matching algorithms such as CoAM \cite{CoAM}, LoFTR \cite{LoFTR}, and
SuperPoint \cite{SuperPoint}, which is the detector and descriptor used in
SuperGlue \cite{SuperGlue}. Hence, we applied SuperGlue \cite{SuperGlue}
scheme using SuperPoint \cite{SuperPoint} and the proposed multisensor
descriptors. We examine the effect of our method on the VIS-NIR dataset \cite%
{SiameseCrossSpectral}. Following similar experiments by Ben Baruch et al.
\cite{multisensor} we define an inlier as a matched pair of keypoints with a
maximal matching error of 5 pixels. We extract the top 200 keypoints
detected by each method, to compute the corresponding descriptors using our
method, and report the mean number of inliers, outliers, and pixel distance
error in Table \ref{tab:image_matching}. Additional qualitative
visualizations are shown in Figure \ref{fig:image_matching}. Our method
improves the SuperPoint \cite{SuperPoint} inliers number by 7\%, and reduces
outliers and distance error by 27\% and 90\%, respectively. Significant
improvement is also shown for CoAM \cite{CoAM}, with a 65\% reduction of the
distance error. LoFTR \cite{LoFTR} is the only method whose gain is less
significant. In particular, comparing the accuracy of all matching schemes
\textit{without} using our approach shows that LoFTR is significantly more
accurate, reducing the matching error by 83\% and 92\%, respectively. Hence,
the use of our more accurate descriptor improves its results but not
significantly. We attribute this to LoFTR's architecture using a Transformer
with a global receptive field.
\begin{table*}[tbh]
\centering
\par
\begin{tabular}{lccc}
\toprule Method & Inliers & Outliers & \makecell{Matching Error\\ $\bigl[
pixels \bigr]$} \\
\midrule SuperPoint\cite{SuperPoint} & 160 & 40 & 140 \\
\textbf{SuperPoint\cite{SuperPoint} + Ours} & \textbf{171} & \textbf{29} &
\textbf{14} \\
CoAM\cite{CoAM} & 189 & 11 & 57 \\
\textbf{CoAM\cite{CoAM} + Ours} & \textbf{190} & \textbf{10} & \textbf{20}
\\
LoFTR\cite{LoFTR} & 192 & 8 & 12 \\
\textbf{LoFTR\cite{LoFTR} + Ours} & 189 & \textbf{7} & \textbf{10} \\
\bottomrule &  &  &
\end{tabular}%
\caption{The matching accuracy of SOTA matching schemes using the proposed
multimodal image descriptor. We compare to applying the matching schemes
with the SuperPoint \protect\cite{SuperPoint} SOTA\ descriptor. We report
the mean number of inliers, outliers, and matching error (in pixels) on the
VIS-NIR dataset \protect\cite{SiameseCrossSpectral}. The inliers are defined
as matched keypoints with a distance of less than 5 pixels \protect\cite%
{multisensor}.}
\label{tab:image_matching}
\end{table*}

\subsection{Attention maps}

\label{subsec:Attention maps}

We further study the Transformer-Encoder self-attention mechanism by
visualizing the attention weights as heatmaps. This allows to analyze the
visual cues the encoder pays attention to. We show the attention heatmaps of
four matching RGB and NIR pairs taken from the VIS-NIR Benchmark \cite%
{SiameseCrossSpectral}, originated from different scenes. The heatmaps are
generated by upsampling the attention weights of the last encoder layer.

\begin{figure}[tbh]
\centering
\begin{tabular}{ccc}
& RGB & NIR \\
\raisebox{5em}{(a)} & \includegraphics[width=0.4%
\linewidth]{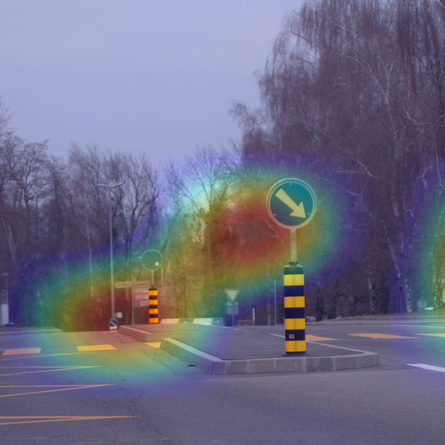} & %
\includegraphics[width=0.4%
\linewidth]{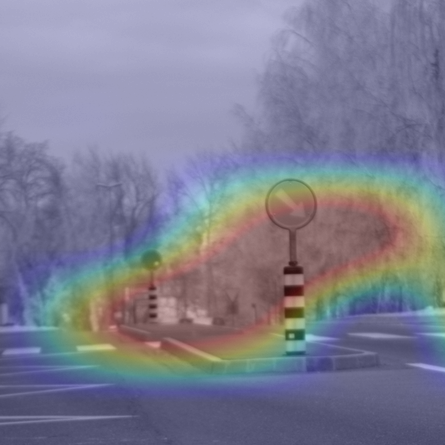} \\
\raisebox{5em}{(b)} & \includegraphics[width=0.4%
\linewidth]{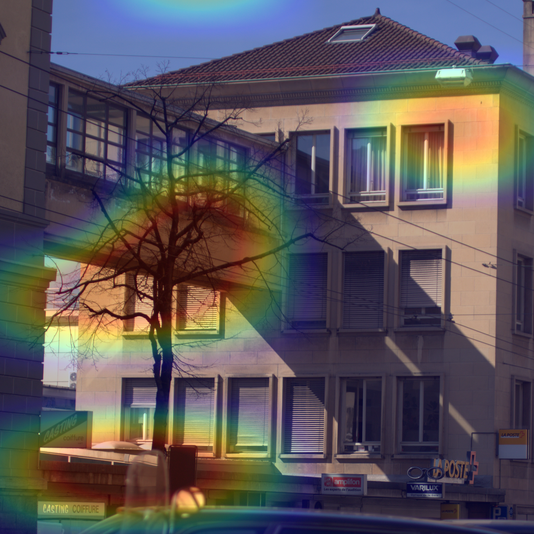} & %
\includegraphics[width=0.4%
\linewidth]{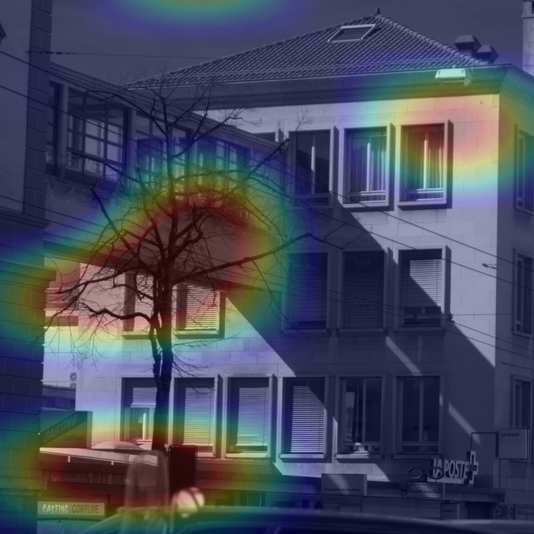} \\
\raisebox{5em}{(c)} & \includegraphics[width=0.4%
\linewidth]{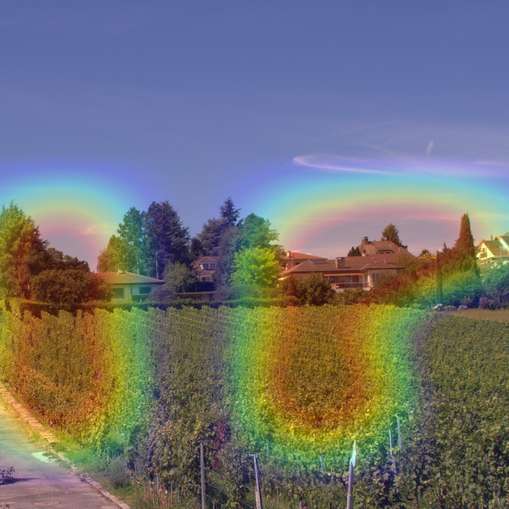} & %
\includegraphics[width=0.4%
\linewidth]{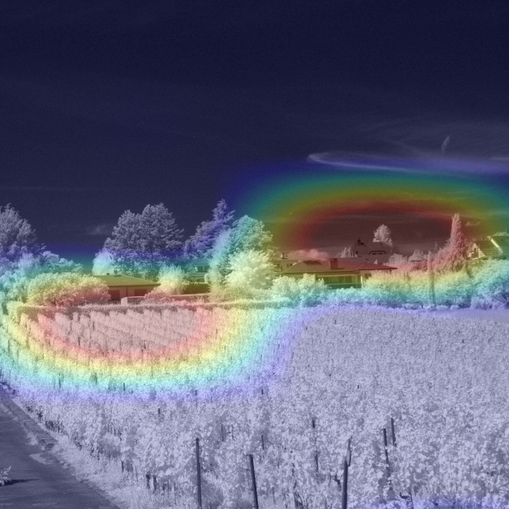} \\
\raisebox{5em}{(d)} & \includegraphics[width=0.4%
\linewidth]{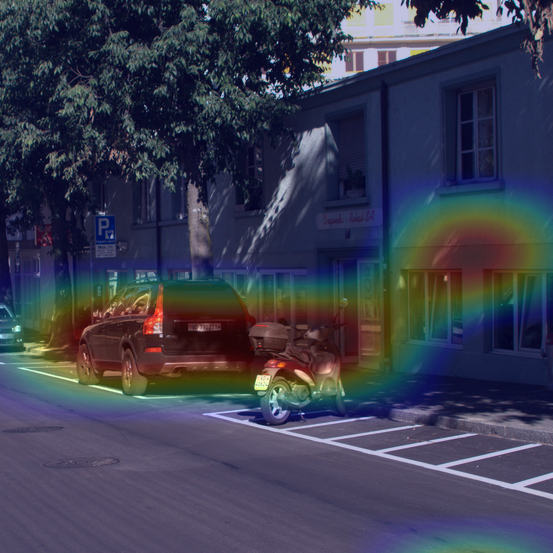} & %
\includegraphics[width=0.4%
\linewidth]{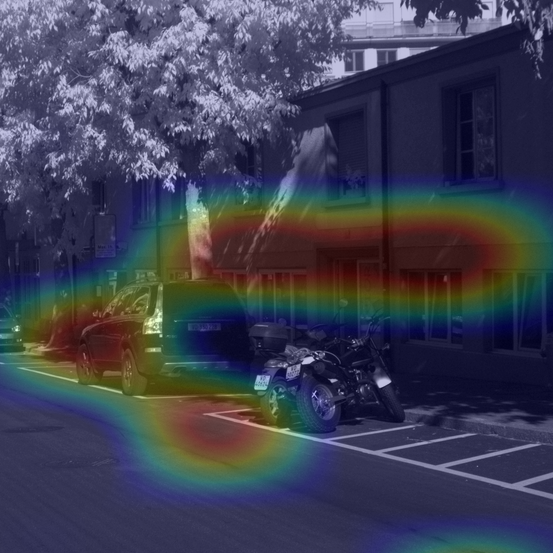} \\
&  &
\end{tabular}%
\caption{Attention heatmaps visualizing the Transformer-Encoder attention
weights. The presented heatmaps show corresponding multimodal RGB and NIR
images taken from various scenes of the VIS-NIR Benchmark \protect\cite%
{SiameseCrossSpectral}.}
\label{fig:attention_heatmaps}
\end{figure}

The resulting attention maps are presented in Fig. \ref%
{fig:attention_heatmaps}, showing that the encoder learns to pay more
attention to locations of modality-invariant features, paying more attention
to locations consisting of object blobs or their distinctive corners and
edges. In the first pair \ref{fig:attention_heatmaps}a, the encoder
summarizes the scene well by paying attention to both the bigger traffic
sign in the middle and the second, smaller one in the background. In the
second pair \ref{fig:attention_heatmaps}b, the encoder attends to the
distinctively shaped tree on the left, as well as the house corner, which is
a good modality-invariant feature. The third pair \ref%
{fig:attention_heatmaps}c contains two small houses, and the encoder chooses
to attend to both, also paying some attention to the grass fields in the
foreground. In the fourth pair \ref{fig:attention_heatmaps}d, the encoder
catches and attends the car in the middle of the scenes along with parts of
the background house, but misses the black motorbike in the NIR image,
probably due to occlusion in the NIR image.

\subsection{Ablation study}

\label{subsec:Ablation study}

\begin{table*}[tbh]
\centering
\begin{tabular}{lcccccc}
\toprule CNN & Transformer & \multicolumn{2}{c}{Attention} & Residual & SPP
& VIS-NIR \\
&  & Layers & Heads &  &  &  \\
\midrule Siamese & - & - & - & - & + & 1.77 \\
Siamese & - & - & - & - & - & 1.83 \\
Siamese & Encoder & 4 & 2 & + & + & 1.5 \\
Siamese & Encoder & 2 & 4 & + & + & 1.59 \\
Siamese & \makecell{Encoder+\\Decoder} & 2 & 2 & + & + & 1.66 \\
\makecell{Pseudo\\Siamese} & Encoder & 2 & 2 & + & + & 2.03 \\
Siamese & Encoder & 2 & 2 & - & + & DIV \\
\textbf{Siamese} & \textbf{Encoder} & \textbf{2} & \textbf{2} & \textbf{+} &
\textbf{+} & \textbf{1.44} \\
\bottomrule &  &  &  &  &  &
\end{tabular}%
\caption{Ablation results evaluated on the VIS-NIR Benchmark \protect\cite%
{SiameseCrossSpectral}. Score is given in terms of FPR95. DIV implies that
the model diverged and was not able to learn.}
\label{table:ablation}
\end{table*}

To better understand the contribution of each component in our proposal, we
experiment with modifying one component or related parameter at a time and
measuring its impact. The experiments were performed on the VIS-NIR
Benchmark \cite{SiameseCrossSpectral}, and are reported at Table \ref%
{table:ablation}. We first evaluate the impact of the Transformer-Encoder.
When the transformer is removed and the rest of the architecture is intact,
it is shown that the addition of the transformer provides a 23\%
improvement. Removing also the SPP as well as the Transformer results in
even a greater error. Performance is not improved by adding encoder layers
or heads, nor by adding a Transformer-Decoder on top of the encoder, which
degrades performance by 14\%, probably due to overfitting. A CNN of unshared
weights, coined Pseudo-Siamese CNN and used in previous SOTA approaches \cite%
{SiameseCrossSpectral, multisensor}, ends up with inferior performance
compared to the Siamese CNN, illustrating the importance of weight sharing.
An additional observation is the significance of the residual connection, as
without it the architecture cannot be trained end-to-end from scratch and
the training diverges.

We further study the SPP layer impact. It is shown in Table \ref%
{table:spp_ablation} that a four-level SPP yields the best results. Three
levels degrade performance, as the $8\times 8$ layer was shown to contain
valuable information for the model, and an additional level also degrades
performance, illustrating that the additional low-resolution layer is
insignificant. A reduced batch size was used due to memory constraints
imposed by the $16\times 16$ level.
\begin{table}[tbh]
\centering
\begin{tabular}{cc}
\toprule Pyramid levels & VIS-NIR \\
\midrule $4\times 4, 2\times 2, 1\times 1$ & 1.75 \\
$8\times 8, 4\times 4, 2\times 2, 1\times 1$ & \textbf{1.6} \\
$16\times 16, 8\times 8, 4\times 4, 2\times 2, 1\times 1$ & 1.65 \\
\bottomrule &
\end{tabular}%
\caption{SPP levels ablation results evaluated on the VIS-NIR Benchmark
\protect\cite{SiameseCrossSpectral}. The score is given in terms of FPR95.
Due to memory constraints, a smaller batch size was used.}
\label{table:spp_ablation}
\end{table}
\begin{table}[tbh]
\centering
{\setlength{\tabcolsep}{3pt} }
\par
{\
\begin{tabular}{ccc}
\toprule Dimensions & Type & VIS-NIR \\
\midrule - & - & 1.63 \\
\multirow{ 2}{*}{1D} & Fixed & 1.54 \\
& Learned & 1.58 \\
\multirow{ 2}{*}{2D} & Fixed & 1.62 \\
& Learned & \textbf{1.44} \\
\bottomrule &  &
\end{tabular}%
}
\caption{Positional encoding ablation results evaluated on the VIS-NIR
Benchmark \protect\cite{SiameseCrossSpectral}. Score is given in terms of
FPR95.}
\label{table:ablation_pos}
\end{table}

We also measure the impact of the different formulations of positional
encoding in Table \ref{table:ablation_pos}, where we evaluated fixed and
learned 1D and 2D positional encodings, following Parmar et al. \cite%
{ImageTransformer} and Eq. \ref{eq:2d_emb}. It follows that learning a 2D
positional encoding yields the best performance, implying that the encoder
benefits from utilizing the original 2D spatial layout of the features.

The embedding dimension used by the network was studied in Table \ref%
{table:ablation_desc} by training the network from scratch using multiple
dimensions. A larger embedding dimension can improve the network's capacity,
but can also lead to overfitting. The results in Table \ref%
{table:ablation_desc} show that $%
\mathbb{R}
^{128}$ is the sweet spot, as the model does not benefit from a larger
descriptor size and its performance is degraded by using a smaller
descriptor.
\begin{table}[tbh]
\centering
\begin{tabular}{cc}
\toprule Dimension & VIS-NIR \\
\midrule 64 & 1.93 \\
\textbf{128} & \textbf{1.44} \\
256 & 1.45 \\
\bottomrule &
\end{tabular}%
\caption{Descriptor size ablation results evaluated on the VIS-NIR Benchmark
\protect\cite{SiameseCrossSpectral}. The score is given in terms of FPR95.}
\label{table:ablation_desc}
\end{table}

\section{Conclusions}

In this paper, we presented a novel approach for performing multimodal image
patch matching using a Transformer-Encoder on top of Siamese CNN multiscale
feature maps, utilizing long-range feature interactions as well as
modality-invariant feature aggregations. We also introduced a residual
connection, which has been shown to be essential for training
Transformer-based networks from scratch. The proposed scheme achieves new
SOTA performance when applied to contemporary multimodal patch matching
benchmarks and the popular single modality UBC Benchmark, illustrating its
generality.


\bibliographystyle{IEEEtran}
\bibliography{Multisensor}

\end{document}